\documentclass[journal]{IEEEtran}
\usepackage{amsmath}
\usepackage{eucal}
\usepackage{amssymb}
\usepackage[ruled,linesnumbered,lined,boxed,commentsnumbered, ruled, vlined]{algorithm2e}
\usepackage{graphicx}
\usepackage{epstopdf}
\graphicspath{{../Painting/}}
\DeclareGraphicsExtensions{.pdf, .jpeg, .png, -eps-converted-to.pdf}

\makeatletter
\newcommand{\removelatexerror}{\let\@latex@error\@gobble}
\makeatother

\usepackage{multirow}
\usepackage{booktabs}
\usepackage{bigstrut}
\usepackage{hyperref}
\usepackage[table]{xcolor}
\usepackage{cite}
\usepackage[square, comma, numbers,sort&compress]{natbib}
\hypersetup{
    colorlinks,
    citecolor=black,
    linkcolor=black,
    urlcolor=black}

\ifCLASSINFOpdf
\else
\fi

\begin{document}
%
\title{DAGs with Tears: A Novel Structure Learning \\ Method under Deep Learning Framework}
%
%
%

\author{Zhichao~Chen,
        and~Zhiqiang~Ge,~\IEEEmembership{Senior~Member,~IEEE}
\thanks{This work was supported in part by the National Natural Science Foundation of China (NSFC) under Grant 61833014 and Grant 61722310,
    and in part by the Natural Science Foundation of Zhejiang Province under Grant LR18F030001. (Corresponding author: Zhiqiang Ge)}
    \thanks{The authors are with the State Key Laboratory of Industrial Control
    Technology, College of Control Science and Engineering, Zhejiang
    University, Hangzhou 310027, China (e-mail: 12032042@zju.edu.cn; gezhiqiang@zju.edu.cn).}
}

%
%

\markboth{Paper Submitted to arXiv}%
{Shell \MakeLowercase{\textit{et al.}}: Bare Demo of IEEEtran.cls for IEEE Journals}
%



\maketitle

\begin{abstract}
    Bayesian network is a frequently-used method for fault detection and diagnosis in industrial processes.
    The basis of Bayesian network is structure learning which learns a directed acyclic graph from data.
    Since the search space in structure learning will scale super-exponentially with the increase of process variables,
    data-driven structure learning is a challenging problem.
    As a novel method for structure learning, DAGs with No Tears methods are being well studied in recent years due to their compatibility with deep learning framework.
    However, the DAGs with No Tears methods is far from application in industrial scenario due to problems in the gradient descent based solving stage and the post-processing stage.
    In this work, those problems are theoretically analyzed in detail by mathematical derivations.
    To solve these problems, the DAGs with Tears method is proposed by using mix-integer linear programming under the deep learning framework.
    In addition, prior knowledge is able to incorporate into the new proposed method, making structure learning more practical and useful in industrial processes.
    Finally, a numerical example and an industrial simulation example are adopted as case studies to demonstrate the superiority of the developed method.
\end{abstract}


%
\IEEEpeerreviewmaketitle

\section{Introduction}
%
%
%
%
\IEEEPARstart{D}{UE} {TO} the high complexity mechanism and the lack of rigorous mathematical models,
the mechanism-driven plant-level optimization is far from widespread application \cite{RN349}, \cite{RN348}.
Thanks to the great improvement of industrial intelligence and information technology,
the acquisition and wide application of industrial big data have become possible.
Therefore, the application of data-driven optimization is being a trending research topic.
Quantities of new data-driven modeling method \cite{RN346}, \cite{RN345} have recently been proposed especially using the deep learning method \cite{RN307},
which play a crucial part in the guidance of control, the increasing of economic value, and the guarantee of process safety.
However, using the data acquired from the process directly could not promise the reliability of the model and results
in the obstacle of the data-driven method promotion \cite{RN320}.
Meanwhile, as a branch of probabilistic graphical model, Bayesian network \cite{RN338} is being more attractive,
which can be regarded as a data structure that provides the skeleton for representing a joint distribution compactly in a factorized way \cite{RN324}.
As a compromise proposal of mechanism and data, Bayesian network makes it possible to open the black box of data-driven models.
The fault detection \& diagnosis technology \cite{RN325}, \cite{RN315} and soft sensor method \cite{RN308} based on Bayesian network have been well studied and hence the research on the application of Bayesian network in industrial big data is of great importance.\\
\indent The construction of Bayesian network consists of parameter learning and structure learning \cite{RN324}.
The structure learning is the basic of Bayesian network construction and a popular research topic in this field \cite{RN330}.
Actually, learning the directed acyclic graph (DAG) from the data directly as the graph of Bayesian network is an NP-Hard problem.
The major difficulty is the combinational explosion of binary variables and the non-convexity for the acyclic constraint in the optimization problem.
To learn the DAGs from data, three major methods are being used namely constraint based method, score based method, and DAGs with No Tears (NOTEAR) \cite{RN314} method.
Constraint based methods like PC algorithm \cite{RN312} view a Bayesian network as a representation of independencies.
They try to test for conditional dependence and independence in the data and then to find a network (or more precisely an equivalence class of networks) that best explains these dependencies and independencies.
Score-based methods like K2 algorithm \cite{RN303}, MCMC algorithm \cite{RN309}, and Hill Climbing Search algorithm \cite{RN339} view a Bayesian network as specifying a statistical model and then address learning as a model selection problem.
Score based methods all operate on the same principle: Define a hypothesis space of potential models — and a scoring function that measures how well the model fits the observed data.
The mentioned above methods are usually based on heuristic rule and could not consider all nodes simultaneously to find an optimal structure.
Different from traditional heuristic methods, GoBNILP algorithm \cite{RN318}, as a score based method, used the mix-integer linear programming (MILP) to traverse all the nodes to maximize the score function.
The prior knowledge can be added in the Bayesian network via the regulation of binary variables, while the combinational optimization and the corresponding constraints makes it to be an NP-Hard problem when solving the MILP linear programming problem.
As a combination of score based and combinational optimization method, NOTEAR utilizes the statistical properties of the least square loss of structural equation model (SEM) in scoring DAGs.
That is, the minimizer of the least square loss in the SEM provably recovers a true DAG with high probability on finite-samples and in high dimensions \cite{RN306}.
Therefore, the combinational optimization of DAG structure learning can be converted into continuous optimization problem and the computational efficiency and performance can be improved substantially.\\
\indent The earliest NOTEAR method is a linear model, which makes it difficult to confront with the strong nonlinearity of process data.
Therein, the least square loss of structural equation model in NOTEAR under deep learning framework is being studied in recent years to improve the accuracy when applying NOTEAR method to nonlinear data.
Yu et al. \cite{RN329} adopted the graph convolution operation to combine the VAE and NOTEAR as well as changed the constraint in NOTEAR simultaneously.
The so-called DAG-GNN model is the first method that extends the NOTEAR to the nonlinear data under deep learning framework and the non-convexity constraint of the original is improved in a power form.
Ng et al. \cite{RN305} analyzed the SEM in an abstract function form and generalized the NOTEAR method to nonlinear case. Wang et al. \cite{RN310} proposed a generative adversarial framework for NOTEAR to improve the DAG mining capability in nonlinear data.\\
\indent Even though the NOTEAR has been well studied and achieved considerable improvement.
In practice, the non-convexity of acyclic constraint is difficult to be satisfied under the deep learning framework whose parameters are updated via gradient descent method \cite{RN347},
which results in the infeasible solution in the final results constantly.
Therein, the DAGs obtained from the NOTEAR methods should make post-processing to satisfy the acyclic constraint.
Generally, the tear problem is converted to be a truncation problem \cite{RN323},
where some elements in the adjacent matrix will be set to 0 to avoid the tear problem.
This post-processing method may have great perturbation of least square result.
Meanwhile, to the best of the our knowledge, the NOTEAR is purely data-driven method,
the prior knowledge cannot be added into the final DAG.
This drawback reduces the reliability of the DAG since the models driven by mechanism and data may be more effective than those purely driven by data.
To solve the two problems mentioned above, the DAGs with Tears (WITHTEAR) method is proposed in this work to take the advantages of NOTEAR and combinational optimization into account.
The innovation of this work can be summarized as follows:\\
\indent 1)	The occurrence of infeasible solution in NOTEAR method is theoretically  analyzed under the gradient descent method,
and the principle of the truncation operation in the post processing stage is analyzed in the perspective of least square perturbation.\\
\indent 2)	The WITHTEAR method is formulated based on the NOTEAR method using MILP model.
Besides, the MILP model considering prior knowledge of the data is combined into the DAG structure learning.\\
\indent The paper is organized as follow: the problem statement is given in Section 2.
In Section 3, the corresponding analyses of NOTEAR are provided in detail, and the WITHTEAR is given in Section 4.
Two case studies are given in Section 5 to show the superiority of WITHTEAR baes on DAG-GNN model.
Conclusions are drawn in Section 6.

\section{Problem Statement}
The problem to be solved in this paper is given as follows: Givens are \emph{n} variables \emph{x}$_i$ which belongs to a set \emph{X} defined as \emph{X}= \{\emph{x}$_1$, \emph{x}$_2$, …, \emph{x}$_n$\}.
The prior knowledge of the variables is presented by the connection between variables using matrix \emph{P}.
The problem is how to combine NOTEAR and MILP model to learn Bayesian Network (DAG) under deep learning framework
and perform the corresponding DAG topological structure using adjacent matrix \emph{A}.

\section{Theoretically analyses of NOTEAR method}
\subsection{The occurrence of infeasible solution}

To better perform the WITHTEAR method, the analysis of NOTEAR method is given as follows.
According to reference \cite{RN314}, the NOTEAR methods convert the traditional combinational optimization problem:
\[\begin{array}{*{20}{c}}
{\mathop {\min }\limits_A {\rm{  }}F(W)}\\
{s.t. {\rm{  }}G(W) \in DAGs}
\end{array}\]
into a continuous programming problem shown as Problem P1:\\
\[(P1){\rm{      }}\begin{array}{*{20}{c}}
{\mathop {\min }\limits_{A,\Theta } \frac{1}{{2n}}\sum\limits_{j = 1}^n {\left\| {{X_{(j)}} - f({X_{(j)}}|A,\Theta )} \right\|_F^2}  + \lambda {{\left\| A \right\|}_1}}\\
{s.t. {\rm{  }}h(A) = 0}
\end{array}\]
\indent The equality constraint can be written as Eq. (1) \cite{RN314} or Eq. (2) \cite{RN329}:

\begin{equation}
    \begin{aligned}
    &&  h(A) = Tr({e^{A \odot A}}) - d = 0
    \end{aligned}
\end{equation}
\begin{equation}
    h(A) = Tr[{(I + \gamma A \odot A)^d}] - d = 0
\end{equation}

where the \emph{Tr} is the trace of matrix, $ \odot $ is Hadmard product, $ \gamma $ is hyper-parameter, \emph{d} is the dimension of square matrix \emph{A}.\\
\indent To solve this constrained optimization problem, according to the reference \cite{RN337}, the augmented Lagrangian method is adopted and the objective function of Problem P1 without constraint can be rewritten as shown in Eq. (3).
The $ \alpha $ and $ \beta $ in Eq. (3) are Lagrange multiplier and penalty parameters respectively. When $\emph{c}\rightarrow \infty$,
the minimizer of Eq. (3) must satisfy $h(A)= 0$, in which case Eq. (3) is equal to the objective function of Problem P1.
\begin{equation}\begin{aligned}
    \mathop {\min }\limits_{A,\Theta } \frac{1}{{2n}}\sum\limits_{j = 1}^n &{\left\| {{X_{(j)}} - f({X_{(j)}}|A,\Theta )} \right\|_F^2}\\  &+ \lambda {\left\| A \right\|_1} + \alpha h(A) + \frac{\beta }{2}{\left| {h(A)} \right|^2}
    \end{aligned}
\end{equation}

\noindent Hence the strategy is progressively increase \emph{c}, and the Lagrange multiplier $ \lambda $ is correspondingly updated as shown in Eq. (4) and (5).

\begin{equation}
{\alpha _{k + 1}} = {\alpha _k} + {\beta _k}h({A_k})
\end{equation}
\begin{equation}
{\beta _{k + 1}} = \left\{ \begin{array}{l}
                                 10 {\beta _k},{\rm{  }}if{\rm{   }}\left| {h({A_k})} \right| > 0.25\left| {h({A_{k - 1}})} \right|{\rm{ }}\\
                                 {\beta _k},{\rm{  }}otherwise
\end{array} \right.\
\end{equation}

%

\indent However, in the iteration process of the gradient descent method, the \emph{A} may not satisfy the acyclic constraint and results in an infeasible solution.
To illustrate this phenomenon, the Problem P1 will be simplified as a linear form as Problem P2 shown.
The constraint written in Eq. (1) and (2) will be discussed respectively.

\[(P2){\rm{      }}\begin{array}{*{20}{c}}
{\mathop {\min }\limits_A \frac{1}{{2n}}\sum\limits_{j = 1}^n {\left\| {{X_{(j)}} - {X_{(j)}}A} \right\|_F^2} }\\
{s.t.{\rm{  }}  h(A) = 0}
\end{array}\]
\indent When Eq. (1) is adopted as constraint, the objective function can be formulated as Eq. (6) shown:
\begin{equation}
\begin{aligned}
    Loss =& \frac{1}{{2n}}\sum\limits_{j = 1}^n {\left\| {{X_{(j)}} - {X_{(j)}}A} \right\|_F^2}  \\
    &+ \alpha (Tr({e^{A \odot A}}) - d) + \frac{\beta }{2}{\left| {Tr({e^{A \odot A}}) - d} \right|^2}
\end{aligned}
\end{equation}

\indent Assume that in the iteration of gradient descent, the \emph{A}$_k$ calculated in the \emph{k}-th time iteration satisfies the constraint, then the \emph{A}$_k$$_+$$_1$ in the next time iteration is shown in Eq. (7).
\begin{equation}
{A_{k + 1}} = {A_k} - LR \times \frac{{\partial Loss}}{{\partial {A_k}}}
\end{equation}
\noindent where \emph{LR} corresponds to the learning rate in the gradient descent method.\\
\indent The Eq. (7) can be expanded \cite{RN334}, \cite{IMM2012-03274} in Eq. (8):
\begin{equation}
\begin{aligned}
{A_{k + 1}} = {A_k} - LR \times & [\frac{1}{n}\sum\limits_{j = 1}^n {X_{(j)}^T({X_{(j)}}{A_k} - {X_{(j)}})} \\
   &  + 2\alpha {A_k} \odot {({e^{{A_k} \odot {A_k}}})^T}]
\end{aligned}
\end{equation}
\noindent then, Eq. (9) and (10) can be derived by multiplying \emph{A}$_k$ and \emph{A}$_k$$_+$$_1$ on both sides respectively.
\begin{equation}
    \begin{aligned}
    {A_{k + 1}} \odot {A_k} = & {A_k} \odot {A_k} \\
    & - LR \times [\frac{1}{n}\sum\limits_{j = 1}^n {X_{(j)}^T({X_{(j)}}{A_k} - {X_{(j)}})} \\
    & + 2\alpha {A_k} \odot {({e^{{A_k} \odot {A_k}}})^T}] \odot {A_k}
\end{aligned}
\end{equation}
\begin{equation}\begin{aligned}
{A_{k + 1}} \odot {A_{k + 1}}  = & {A_k} \odot {A_{k + 1}} \\
   & - LR \times [\frac{1}{n}\sum\limits_{j = 1}^n {X_{(j)}^T({X_{(j)}}{A_k} - {X_{(j)}})} \\
   & + 2 \alpha {A_k} \odot {({e^{{A_k} \odot {A_k}}})^T}] \odot {A_{k + 1}}
\end{aligned}
\end{equation}
\noindent Substitute Eq. (10) to Eq. (9), Eq. (11) can be given as follow:
\begin{equation}\begin{aligned}
{A_k} \odot {A_k} = &  {A_{k + 1}} \odot {A_{k + 1}} \\
& + LR \times [\frac{1}{n}\sum\limits_{j = 1}^n {X_{(j)}^T({X_{(j)}}{A_k} - {X_{(j)}})} \\
& + 2\alpha {A_k} \odot {({e^{{A_k} \odot {A_k}}})^T}] \odot ({A_{k + 1}} + {A_k})
\end{aligned}
\end{equation}
\noindent Finally, the LHS of Eq. (11) is part of constraint Eq. (1), Hence, Eq. (12) can be derived when substituting Eq. (11) to Eq. (1):
\begin{equation}
\begin{aligned}
    h\{& {A_{k + 1}} \odot {A_{k + 1}} \\
    & + LR \times [\frac{1}{n}\sum\limits_{j = 1}^n {X_{(j)}^T({X_{(j)}}{A_k} - {X_{(j)}})} \\
    & + 2\alpha {A_k} \odot {({e^{{A_k} \odot {A_k}}})^T}] \odot ({A_{k + 1}} + {A_k})\}  = 0
\end{aligned}
\end{equation}
\indent Since the data is fed batch-by-batch, and the direction of gradient descent may not be determined due to the data shuffling every time of iteration.
Besides, the \emph{LR} as hyper-parameters could not be appropriate to ensure the \emph{A} be feasible at the last time of iteration.
Therein, Eq. (13) can’t be permanent establishment during iteration of Eq. (12), which means that the algorithms will report an infeasible solution of \emph{A} at the end of iteration.
\begin{equation}
    \begin{aligned}
        & [ {\frac{1}{n}\sum\limits_{j = 1}^n {X_{(j)}^T({X_{(j)}}{A_k} - {X_{(j)}})}  + 2\alpha {A_k} \odot {{({e^{{A_k} \odot {A_k}}})}^T}}] \\
        & \odot ({A_{k + 1}} + {A_k}) = 0
    \end{aligned}
\end{equation}

\indent Next, Eq. (2) is adopted in Problem P2, and the objective function can be written as Eq. (14):
\begin{equation}
    \begin{aligned}
        Loss =  & \frac{1}{{2n}}\sum\limits_{j = 1}^n {\left\| {{X_{(j)}} - {X_{(j)}}A} \right\|_F^2} \\
        & + \alpha \{ Tr[{(1 + \gamma A \odot A)^m}] - d\} \\
        & + \frac{\beta }{2}{\left| {Tr[{{(1 + \gamma A \odot A)}^m}] - d} \right|^2}
        \end{aligned}
\end{equation}
\noindent Similarly, the \emph{A}$_k$$_+$$_1$ can be derived as shown in Eq. (15):
\begin{equation}
\begin{aligned}
{A_{k + 1}} = & {A_k} - LR \times \{ \frac{1}{n}\sum\limits_{j = 1}^n {X_{(j)}^T({X_{(j)}}{A_k} - {X_{(j)}})} \\
 & + 2\alpha {A_k} \odot \sum\limits_{k = 1}^d {C_d^kk{{\left[ {\gamma {{({A_k} \odot {A_k})}^{k - 1}}} \right]}^T}} \}
\end{aligned}
\end{equation}
\noindent where the $C$ stands for the combinationtorial number and after suitable transformation, Eq. (15) can be rewritten as Eq. (16):
\begin{equation}
    \begin{aligned}
    & {A_k} \odot {A_k} = {A_{k + 1}} \odot {A_{k + 1}} \\
    & + LR \times \{ \frac{1}{n}\sum\limits_{j = 1}^n {X_{(j)}^T({X_{(j)}}{A_k} - {X_{(j)}})} \\
    & + 2\alpha {A_k} \odot \sum\limits_{k = 1}^d {C_d^kk{{\left[ {\gamma {{({A_k} \odot {A_k})}^{k-1}}} \right]}^{T}}} \} \\
    &  \odot ({A_k} + {A_{k + 1}})
        \end{aligned}
\end{equation}
\noindent Substituting Eq. (16) to Eq. (2), the Eq. (17) similar to Eq. (12) can be derived which could not promise a feasible solution after iteration.
\begin{equation}
    \begin{aligned}
        & h[{A_{k + 1}} \odot {A_{k + 1}}  \\
        & + LR \times \{ \frac{1}{n}\sum\limits_{j = 1}^n {X_{(j)}^T({X_{(j)}}{A_k} - {X_{(j)}})} \\
        & + 2\alpha {A_k} \odot \sum\limits_{k = 1}^d {C_d^kk{{\left[ {\gamma {{({A_k} \odot {A_k})}^{k-1}}} \right]}^T}} \} \\
        &  \odot ({A_k} + {A_{k + 1}})] = 0
        \end{aligned}
    \end{equation}
\indent Last, the nonlinear form is adopted in Problem P1 and the similar equations can be given in Eq. (18) and (19), which the similar phenomena can be derived.
\begin{equation}
    \begin{aligned}
        & h\{ {A_{k + 1}} \odot {A_{k + 1}} + LR \\
        &  \times [\frac{1}{n}\sum\limits_{j = 1}^n {{{(\frac{{\partial f({X_{(j)}}|{A_k},\Theta )}}{{\partial {A_k}}})}^T}(f({X_{(j)}}|{A_k},\Theta ) - X_{(j)})} \\
        & + 2\alpha {A_k} \odot {({e^{{A_k} \odot {A_k}}})^T}] \\
        & \odot ({A_{k + 1}} + {A_k}) \} = 0
        \end{aligned}
    \end{equation}
\begin{equation}
    \begin{aligned}
        & h\{ {A_{k + 1}} \odot {A_{k + 1}} + LR  \\
        & \times [\frac{1}{n}\sum\limits_{j = 1}^n {{{(\frac{{\partial f({X_{(j)}}|{A_k},\Theta )}}{{\partial {A_k}}})}^T}(f({X_{(j)}}|{A_k},\Theta ) - X_{(j)})} \\
        &  + 2\alpha {A_k} \odot \sum\limits_{k = 1}^d {C_d^kk{{\left[ {\gamma {{({A_k} \odot {A_k})}^{k-1}}} \right]}^T}} ] \\
        &  \odot ({A_{k + 1}} + {A_k})\} = 0
        \end{aligned}
    \end{equation}
\indent It should be noticed that, during the practice, if the L1 norm regularization is added in the objective function,
the \emph{A} will tends to be 0. The reason can be interpretated through observing Eq. (12), (17), (18) and (19).
With the increase of $\gamma$ as the iterations increases, the existence of the factor shown in the LHS of Eq. (20) gained from the above-mentioned equations tends to be 0,
meanwhile, the L1 norm will forced the \emph{A} to be sparsity \cite{RN335}, and hence the \emph{A} will tend to be 0 in the iteration.
\begin{equation}
{A_{k + 1}} + {A_k} = 0
    \end{equation}
\indent In conclusion, due to the limit of gradient descent method and the non-convexity of the equation constraint,
the DAGs acquired via NOTEAR methods will not promise to be acyclic.
Therein, the post-processing operation accompanied by the NOTEAR methods arises.
In the next section, the principle of post-processing operation (also known as truncate as mentioned before) will be stated for better introducing the WITHTEAR method.
\subsection{The principle of the post-processing operation}
\indent In post-processing operation, a threshold will be set and the elements in matrix \emph{A} lower than this threshold will be set to 0, and the new matrix \emph{A} is obtained.
If the new matrix \emph{A} satisfies the acyclic condition, then the new matrix \emph{A} will be output as the final results, otherwise the new matrix \emph{A} is cyclic,
then the threshold will increase and the matrix will be truncated again until it is acyclic.
In the truncate process, the elements of \emph{A} will make the optimal least square result deviate from the optimal point.
Therefore, the rationality of the truncation operation should be analyzed, which to the best of our knowledge has not been studied before.
A simple analysis of the least square will be provided as follows to show the principle of the post-processing operation.\\
\indent Firstly, the disturbance of the linear least square problem will be considered. According to the least square problem shown as Eq. (21):
\begin{equation}
    \tilde X = XA
\end{equation}
\noindent If a perturbation $\delta$\emph{A} is added to \emph{A}, the regression perturbation on the LHS of Eq. (21) can be given as follow:
\begin{equation}
    \delta \tilde X = X \times \delta A
\end{equation}
\noindent Then, the inequality can be derived:
\begin{equation}
{\left\| {\delta \tilde X} \right\|_2} \leqslant {\left\| {\delta A} \right\|_2}{\left\| X \right\|_2}
\end{equation}
\indent Both sides of Eq. (23) can be divided by the L2 norm of matrix \emph{X} and the relative error of the perturbation can be given as Eq. (24).
Similarly, for the non-linear form, the perturbation is given as Eq. (25).
\begin{equation}
    \begin{aligned}
\frac{{{{\left\| {\delta \tilde X} \right\|}_2}}}{{{{\left\| {\tilde X} \right\|}_2}}} & \le \frac{{{{\left\| A \right\|}_2}{{\left\| X \right\|}_2}}}{{{{\left\| {XA} \right\|}_2}}}\frac{{{{\left\| {\delta A} \right\|}_2}}}{{{{\left\| A \right\|}_2}}}\\
& = (\frac{{{{\left\| A \right\|}_2}{{\left\| X \right\|}_2}}}{{{{\left\| {XA} \right\|}_2} \times {{\left\| A \right\|}_2}}}) \times {\left\| {\delta A} \right\|_2}
\end{aligned}
    \end{equation}
\begin{equation}
\frac{{{{\left\| {\delta \tilde X} \right\|}_2}}}{{{{\left\| {\tilde X} \right\|}_2}}} \le \frac{{{{\left\| {\frac{{\partial f(X|A,\Theta )}}{{\partial A}}} \right\|}_2}}}{{{{\left\| {f{{(X|A,\Theta )}}} \right\|}_2}}}{\left\| {\delta A} \right\|_2}
\end{equation}
\indent Summarizing Eq. (24) and (25), it can be concluded that, once the matrix \emph{A} and paramters of regression model is determined,
the relative error merely depends on the perturbation magnitude, which means that, to make the deterioration of the least square result as less as possible,
the elements of \emph{A} will be set to be 0 from small to large until the graph constructed by \emph{A} is acyclic. This process can adopt tear operation to fulfill.
However, for convenience, the NOTEAR, makes the tear problem into a truncation problem as mentioned before.
Note that, the truncation operation may deteriorate the least square loss greater than that of the tear operation.
Meanwhile, the prior knowledge can’t merge with the knowledge learned from data in NOTEAR through roughly truncation.
Therefore, the following WITHTEAR method will be proposed to solve the two problems mentioned above.

\section{DAGs with Tears Method}
\subsection{Loops Tear by MILP problem}
\indent In the previous section, to promise the minimum perturbation of the least square problem and tear all loops of the graph simultaneously,
the number of elements in \emph{A} should be changed as few as possible.
Therein, MILP model can be formulated to fulfill the two goals as stated above and the corresponding method is named DAGs with Tears (WITHTEAR).
Before showing the WITHTEAR method, the following concept will be stated as follow for a better understanding of WITHTEAR method.\\
\indent Loop matrix given in Eq. (26) is a matrix with connection between node (also known as stream, whose set is written as STR)
as column and loop $i$ as row, $\left[ {{u_{i,j}}} \right]$. If a loop $i$ includes a stream $j$,
the element in the loop matrix $ {{u_{i,j}}} = 1$, otherwise $ {{u_{i,j}}} = 0$.
\begin{equation}
    U = \left[ {\begin{array}{*{20}{c}}
    {\begin{array}{*{20}{c}}
         0\\
         \vdots \\
         1
    \end{array}}&{\begin{array}{*{20}{c}}
                      \ldots \\
                      \ddots \\
                      \ldots
    \end{array}}&{\begin{array}{*{20}{c}}
                      1\\
                      \vdots \\
                      0
    \end{array}}
    \end{array}} \right]
    \end{equation}
\indent Based on Eq. (26), innovated by \cite{RN326}, \cite{RN316}, the loop tearing cost is introduced as Eq. (27) to measure the cost of breaking the stream.
\begin{equation}
    Cost = \sum\limits_{j \in STR} {{w_j} \times {y_j}} ,{\rm{  }}{y_j} \in \{ 0,1\}
\end{equation}
\noindent where \emph{w}$_j$ is the weight of the stream, which can be converted from the coefficient of \emph{A}, and \emph{y}$_j$ is binary variable to be solved.
If \emph{y}$_j$ is 0, then the stream should be tear, vice versa.
To tear all loops, the corresponding constraints is given in Eq. (28).
This constraint indicates that the loop in matrix \emph{U} should be tore at least.
By adopting this constraint, the acyclic of the graph formulated from matrix \emph{A} can be promised.
\begin{equation}
    \sum\limits_{j \in STR} {{u_{ij}}{y_j}}  \ge 1,{\rm{  }}u \in U,{y_j} \in \{ 0,1\}
    \end{equation}
\indent In all, the MILP problem can be given in problem P3.
By solving problem P3,
the loops of matrix \emph{A} can be tore and the final DAG can be form with the minimum deterioration of least square result.\\
\begin{center}
$(P3){\rm{      }}\begin{array}{*{20}{c}}
{\min {\rm{  }}Cost = \sum\limits_{j \in STR} {{w_j} \times {y_j}} }\\
{s.t.{\rm{  }}\left\{ {\begin{array}{*{20}{c}}
{\sum\limits_{j \in STR} {{u_{ij}}{y_j}}  \ge 1,{\rm{  }}u \in U}\\
{{y_j} \in \{ 0,1\} }
\end{array}} \right.}
\end{array}$
\end{center}
\subsection{Loops Tear combing prior knowledge and MILP problem}
\indent Note that, the prior knowledge is not added in the MILP problem P3. Hence, in this section, Problem P3 will be extended to combine prior knowledge with the addition of logical propositions and the corresponding disjunctions.
The prior knowledge can be given in 3 scenarios:(1), the existence of stream \emph{j} is unknown; (2), the existence of stream \emph{j} is obligatory; (3), the existence of stream \emph{j} is forbidden.
\indent For scenario (1), the corresponding elements in matrix A can be set to be 0 before the solving of MILP problem P3. Therein, the scenario (2) and (3) will be discussed as follow:
To show the relative position relationships, the logical variables \cite{RN317} are defined as below:\\
\indent 1). \emph{V}$_1$$_,$$_j$ If the existence of stream \emph{j} is unknown, \emph{V}$_1$$_,$$_j$ is True.\\
\indent 2). \emph{V}$_2$$_,$$_j$ If the existence of stream \emph{j} is obligatory, \emph{V}$_2$$_,$$_j$ is False.\\
\indent The disjunctions can be given as Eq. (29) shown:
\begin{equation}
    \begin{gathered}
        \left[ {\begin{array}{*{20}{c}}
        {{V_{1,j}}} \\
        {U{B_{1,j}} = 1.0} \\
        {L{B_{1,j}} = 0.0} \\
        {j \in STR}
        \end{array}} \right] \vee \left[ {\begin{array}{*{20}{c}}
        {{V_{2,j}}} \\
        {U{B_{2,j}} = 0.5} \\
        {L{B_{2,j}} = 0.0} \\
        {j \in STR}
        \end{array}} \right] \hfill \\
        {V_{1,j}},{V_{2,j}} \in \{ True,False\}  \hfill \\
    \end{gathered}
    \end{equation}
\noindent where \emph{UB} and \emph{LB} are the upper bound and lower bound of binary variable \emph{y}$_j$ defined in Eq. (28).
The extra constraint for the binary variable \emph{y}$_j$ is given as Eq. (30):
\begin{equation}
    L{B_j} \le {y_j} \le U{B_j}
    \end{equation}
\noindent there are two constraints in Eq. (29), while only one constraint will take effect, and the remains do not work.
Therefore, each logical variable should satisfy the constraint shown in Eq. (30):
\begin{equation}
{V_{1,j}}\underset{\raise0.3em\hbox{$\smash{\scriptscriptstyle-}$}}{ \vee } {V_{2,j}}{\text{ }}
    \end{equation}

\indent Finally, the optimization problem can be formulated as problem P4, and the value of the logical variable can be derived from matrix \emph{P} that contains prior knowledge.
\[(P4){\text{  }}\begin{array}{*{20}{c}}
{\min {\text{  }}Cost = \sum\limits_{j \in STR} {{w_j} \times {y_j}} } \\
{s.t.\left\{ \begin{gathered}
                 \sum\limits_{j \in STR} {{u_{ij}}{y_j}}  \geqslant 1,{\text{  }}u \in U \hfill \\
                 {y_j} \in \{ 0,1\}  \hfill \\
                 L{B_j} \leqslant {y_j} \leqslant U{B_j},{\text{  }}j \in STR \hfill \\
                 \left[ {\begin{array}{*{20}{c}}
                 {{V_{1,j}}} \\
                 {U{B_{1,j}} = 1.0} \\
                 {L{B_{1,j}} = 0.0} \\
                 {j \in STR{\text{ }}}
                 \end{array}} \right] \vee \left[ {\begin{array}{*{20}{c}}
                 {{V_{2,j}}} \\
                 {U{B_{2,j}} = 0.5} \\
                 {L{B_{2,j}} = 0.0} \\
                 {j \in STR{\text{  }}}
                 \end{array}} \right] \hfill \\
                 {V_{1,j}},{V_{2,j}} \in \{ True,False\}  \hfill \\
                 {V_{1,j}}\underset{\raise0.3em\hbox{$\smash{\scriptscriptstyle-}$}}{ \vee } {V_{2,j}} \hfill \\
\end{gathered}  \right.}
\end{array}\]

\subsection{The algorithm for DAGs with Tears method}

    \begin{figure}[htbp]
        \removelatexerror
        \begin{algorithm}[H]

            \caption{Pseudo-code of WITHTEAR Method}\label{algorithm}


            \SetKwData{Left}{left}\SetKwData{This}{this}\SetKwData{Up}{up}
            \SetKwFunction{Union}{Union}\SetKwFunction{FindCompress}{FindCompress}
            \SetKwInOut{Input}{input}\SetKwInOut{Output}{output}

            \BlankLine

            \BlankLine
            \Input{Data \emph{X}, Parameter $\beta$$_{max}$, the maximum iteration time of deep learning model \emph{Epoch},
                the initial value of $\alpha$, $\beta$, and matrix \emph{A}. The coefficient of L1 norm $\lambda$.
                The initial value of the best objective function of Problem P1 \emph{Loss}$_{best}$ = $\infty$,
                and matrix \emph{A} \emph{A}$_{best}$ = None. The matrix which loads the prior knowledge \emph{P}. The hyper-parameter $\omega$}
            \Output{The best matrix \emph{A}$_{best}$}
            \BlankLine
            \textbf{Training Stage}\:

            \While{\begin{small}$\beta  \leqslant {\beta _{max}}$\end{small}}{

                \For{i $=$ $1 :$ Epoch}{\begin{small} \[{A_{k + 1}} = \begin{array}{*{20}{c}}
                {\mathop {\arg \min }\limits_A \frac{1}{{2n}}\sum\limits_{j = 1}^n {\left\| {{X_{(j)}} - f({X_{(j)}}|{A_k},\Theta )} \right\|_F^2} } \\
                { + \lambda {{\left\| A \right\|}_1} + \alpha h({A_k}) + \frac{\beta }{2}{{\left| {h({A_k})} \right|}^2}}
                \end{array}\]
                \end{small}
                    \If{\begin{small} $\frac{{\sum\limits_{j = 1}^n {\left\| {{X_{(j)}} - f({X_{(j)}}|{A_k},\Theta )} \right\|_F^2} }}{{2n}} \leqslant Los{s_{best}} $
                    \end{small}}{
                        \begin{small}
                            $Los{s_{best}} = \frac{{\sum\limits_{j = 1}^n {\left\| {{X_{(j)}} - f({X_{(j)}}|{A_k},\Theta )} \right\|_F^2} }}{{2n}}$
                        \end{small} \;
                        \begin{small}
                            ${A_{best}} = {A_k}$
                        \end{small}\;
                    }
                }
                \begin{small}
                    $ {\alpha _{k + 1}} = {\alpha _k} + {\beta _k}h({A_k}) $ \;
                \end{small}
                \begin{small}
                    $ {\beta _{k + 1}} = \left\{ \begin{gathered}
                                                     10 \times {\beta _k},{\text{  }}if{\text{   }}\left| {h({A_k})} \right| > 0.25\left| {h({A_{k - 1}})} \right|{\text{ }} \hfill \\
                                                     {\beta _k},{\text{  }}otherwise \hfill \\
                    \end{gathered}  \right.$ \end{small}\;

            }
        \end{algorithm}
    \end{figure}
\indent According to Problem P3 and P4, the pseudo-code of WITHTEAR method can be given based on the results of NOTEAR as Algorithm 1.
From the pseudo-code, it can be concluded that the training stage has no differences between NOTEAR method,
any NOTEAR under deep learning framework can be used to acquire matrix \emph{A}.\\
\indent In the tear stage, the post processing strategy is changed comparing to the truncation operation adopted in NOTEAR method.
By solving the MILP problem, the prior knowledge of data can be added into the DAGs structure learning increasing the validity of the DAG structure mined from data.
Meanwhile, comparing to traditional DAG structure learning using MILP like GoBNILP,
the binary variables in this research merely depend on the number of streams rather than that of the parent node,
results in a reduction in computation complexity.

\begin{figure}[htbp]
        \removelatexerror

        \begin{algorithm}[H]
            \LinesNumbered
            \setcounter{AlgoLine}{8}

            \BlankLine
            \textbf{Tear Stage}\:
            \BlankLine
            \textbf{\quad Preprocessing }\:
            For the elements in A lower than the hyper-parameter $\omega$ and the existence of corresponding streams are forbidden in \emph{P},
            the 0 should be placed in the corresponding location.
            For those should streams be existence in \emph{P} while the corresponding elements are 0 in \emph{A},
            the corresponding elements set to be \begin{small} ${\left\| A \right\|_{\max }}$ \end{small} \;
            \While{Loops exist in matrix \emph{A}$_{best}$}{
                Formulate loop matrix \emph{U}\;
                Solve Problem $(P3)$ or $(P4)$\;
                For those streams should be tore, set the corresponding elements in \emph{A}$_{best}$ to 0\;
                Detect the existence of loop in A }

        \end{algorithm}
    \end{figure}

\section{Case Studies}
\subsection{Numerical example}
\indent In this section, the DAG-GNN without L1 norm will be adopted to test the effectiveness WITHTEAR method.
More Detailed of DAG-GNN model [20] are listed in the appendix.
The nonlinear data simulation shown in Eq. (31) from [13] is modified as the numerical example.
\begin{equation}
    \begin{aligned}
        X & = \tanh (X{W^T}) + \cos (X{W^T}) + \sin (X{W^T}) \\
        &  + z,z \sim {\mathcal N}(0,I)
    \end{aligned}
\end{equation}
\indent The matrix \emph{W} is upper triangular matrix, and the prior knowledge is the existence of streams connected by the nodes in lower triangular place of matrix \emph{A} is forbidden.
A total of 5, 000 data are used for the structure learning.\\
\indent The score functions are false discovery rate (FDR), true positive rate (TPR),
false positive rate (FPR), and structure hamming distance (SHD).
The corresponding expression are given in Eq. (32) $\sim $ (35).
\begin{equation}
    FDR = \frac{{R + FP}}{{TEE}}
\end{equation}
\begin{equation}
    TPR = \frac{{TP}}{T}
\end{equation}
\begin{equation}
    FPR = \frac{{R + FP}}{F}
\end{equation}
\begin{equation}
    SHD = E + M + R
\end{equation}
\noindent In Eq. (32), the \emph{R} and \emph{FP} are edges predicted reversed and not in the undirected skeleton of the true graph, respectively.
The \emph{TEE} is the total number of estimated edges. In Eq. (33), \emph{TP} and \emph{T} are correct direction edges and true edges respectively.
The \emph{F} is the non-edges in the ground truth graph in Eq. (34).
The \emph{E} is the extra edges from the skeleton while the \emph{M} is the missing edges from the skeleton.\\
\begin{figure}[htb]
    \centering
    \includegraphics[width=7.5cm,height=6cm]{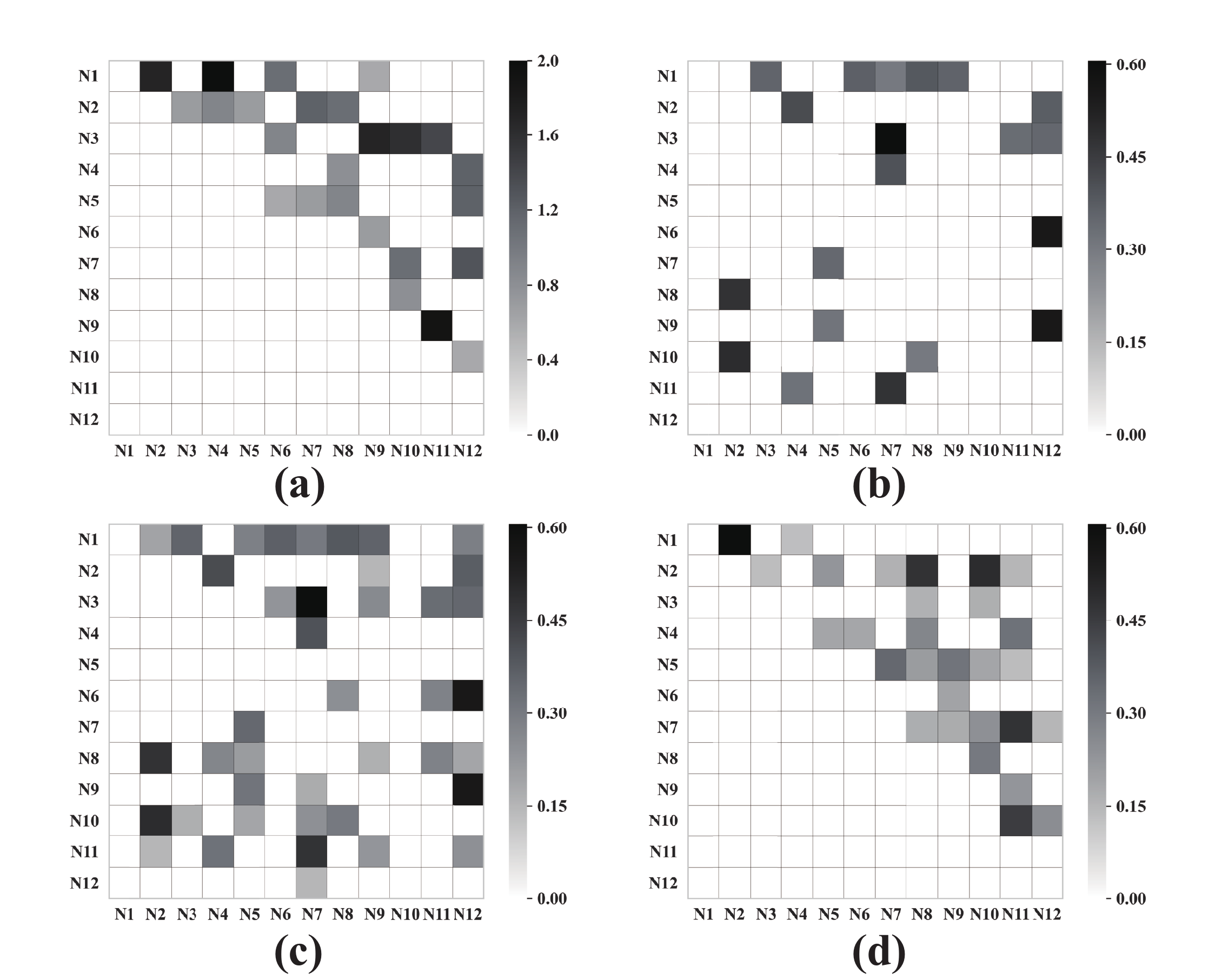}\\
    \caption{The adjacent matrix (a), ground truth; (b), DAG-GNN; (c), DAGs with Tears (Problem P3); (d), DAGs with Tears (Problem P4). }
\end{figure}
\indent Fig. 1 proposed the predicting results and table 2 indicates the score function corresponds to the graphs in Fig. 1.
By using WITHTEAR method, the \emph{FDR}, \emph{SHD} and \emph{FPR} are 5.1$\sim $24.0\%, 14.6$\sim $46.3\%, 46.9$\sim $59.4\% lower than that of the DAG-GNN, respectively.
If the prior is knowledge is not given in WITHTEAR, the \emph{TPR} will 66.67 \% lower than that of the DAG-GNN.
While the situation is changed thanks to the prior knowledge matrix \emph{P} is adopted in the WITHTEAR and results in the \emph{TPR} 77.78\% higher than that of the DAG-GNN.
It can be seen that the prior knowledge on the coefficient matrix \emph{W} can help the elimination of the false edge so as to increase the \emph{TPR}.

\begin{table}[htbp]
    \centering
    \caption{The score functions of the corresponding methods}
    \begin{tabular}{p{6.5em}cccc}
        \toprule
        \multicolumn{1}{c}{ID} & FDR   & TPR   & SHD   & FPR \\
        \midrule
        \multicolumn{1}{c}{DAG-GNN} & 0.81  & 0.36  & 41    & 0.78 \\
        WITHTEARS(P3) & 0.77  & 0.12  & 35    & 0.41 \\
        WITHTEARS(P4) & 0.62  & 0.64  & 22    & 0.32 \\
        \bottomrule
    \end{tabular}%
    \label{tab:addlabel}%
\end{table}%

\subsection{Tennessee-Eastman Process}

\begin{figure}[htb]
    \centering
    \includegraphics[width=9cm,height=6.428571cm]{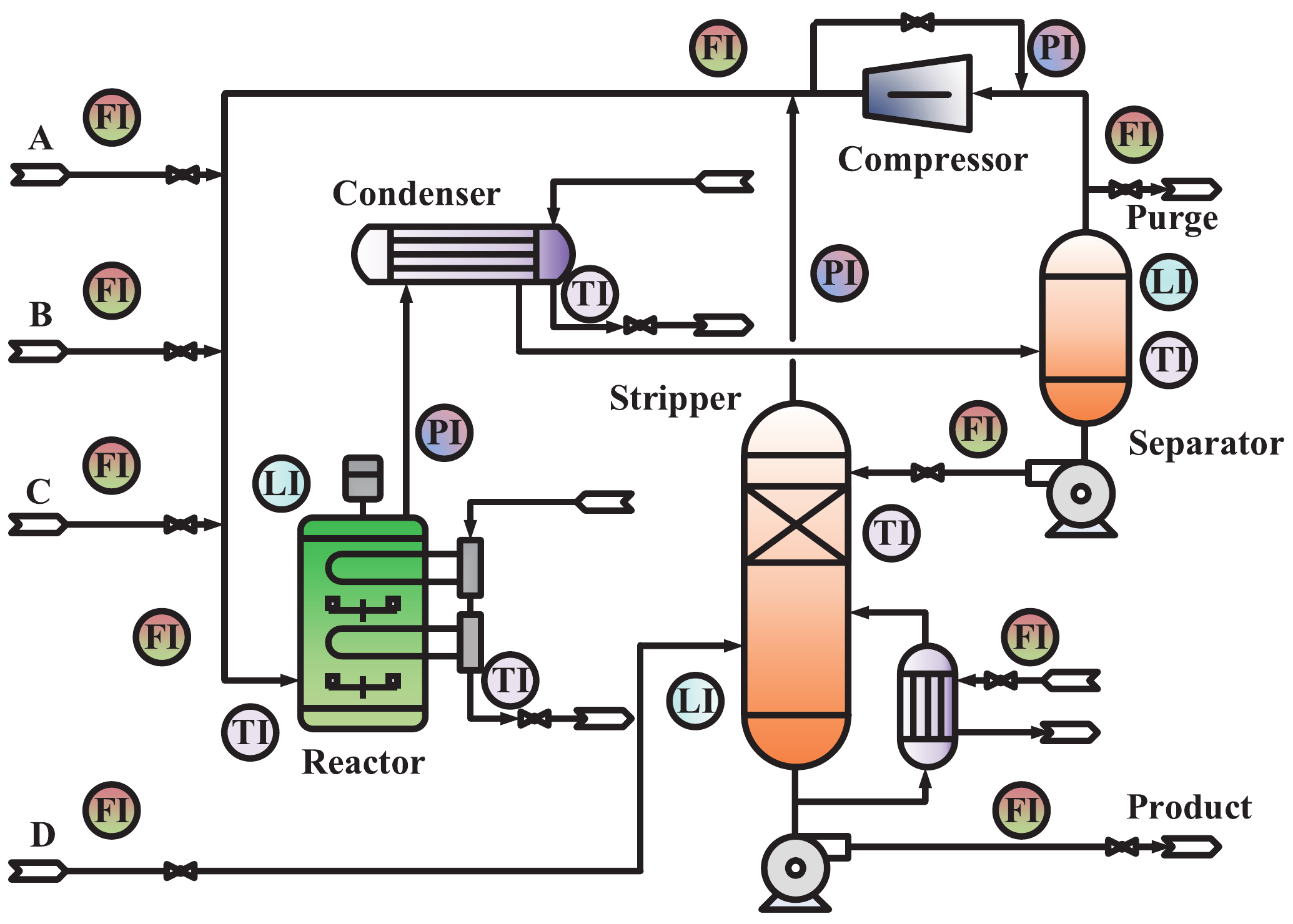}\\
    \caption{The flowsheet of Tennessee-Eastman process. }
\end{figure}
\indent As shown in Fig. 2, the Tennessee Eastman (TE) process has been widely used as a benchmark process \cite{RN333} to compare evaluated process monitoring and fault diagnosis algorithms.
This process includes five major units: a condenser, a reactor, a stripper, a recycle compressor, and a vapor/liquid separator.
There are 41 measured variables, 12 manipulated variables. In this study, 33 variables listed in Table 3 are selected to demonstrate the effectiveness of the DAGs with Tears method.
The dataset used for modeling contains a total of 1, 460 data.


\begin{table}[htb]
    \centering
    \caption{The variables to learn DAG structure in TE process}
    \begin{tabular}{@{}llll@{}}
        \toprule
        No. & Measured Variables            & No. & Measured Variables                                                                   \\ \midrule
        N1  & A feed                        & N18 & Stripper temperature                                                                 \\
        N2  & D feed                        & N19 & Stripper steam flow                                                                  \\
        N3  & E feed                        & N20 & Compressor work                                                                      \\
        N4  & Total feed                    & N21 & \begin{tabular}[c]{@{}l@{}}Reactor cooling water\\ outlet temperature\end{tabular}   \\
        N5  & Recycle flow                  & N22 & \begin{tabular}[c]{@{}l@{}}Separator cooling water\\ outlet temperature\end{tabular} \\
        N6  & Reactor feed rate             & N23 & D feed flow                                                                          \\
        N7  & Reactor Pressure              & N24 & E feed flow                                                                          \\
        N8  & Reactor level                 & N25 & A feed flow                                                                          \\
        N9  & Reactor temperature           & N26 & Total feed flow                                                                      \\
        N10 & Purge rate                    & N27 & Compressor recycle valve                                                             \\
        N11 & Product separator temperature & N28 & Purge valve                                                                          \\
        N12 & Separator level               & N29 & Separator product liquid flow                                                        \\
        N13 & Separator pressure            & N30 & Stripper product liquid flow                                                         \\
        N14 & Separator underflow           & N31 & Stripper steam valve                                                                 \\
        N15 & Stripper level                & N32 & Reactor cooling water flow                                                           \\
        N16 & Stripper pressure             & N33 & Condenser cooling water flow                                                         \\
        N17 & Stripper underflow            &     &                                                                                      \\ \bottomrule
    \end{tabular}
\end{table}
\indent Due to the ground-truth DAG structure is unknown in TE process, the BGe score \cite{RN340} and Gaussian BIC score \cite{RN341} are adopted to score the DAGs \cite{RN350}.
The matrix P which contains prior knowledge is given in Fig. 3.
\begin{figure}[h]
    \centering
    \includegraphics[width=5.8cm,height=5.8cm]{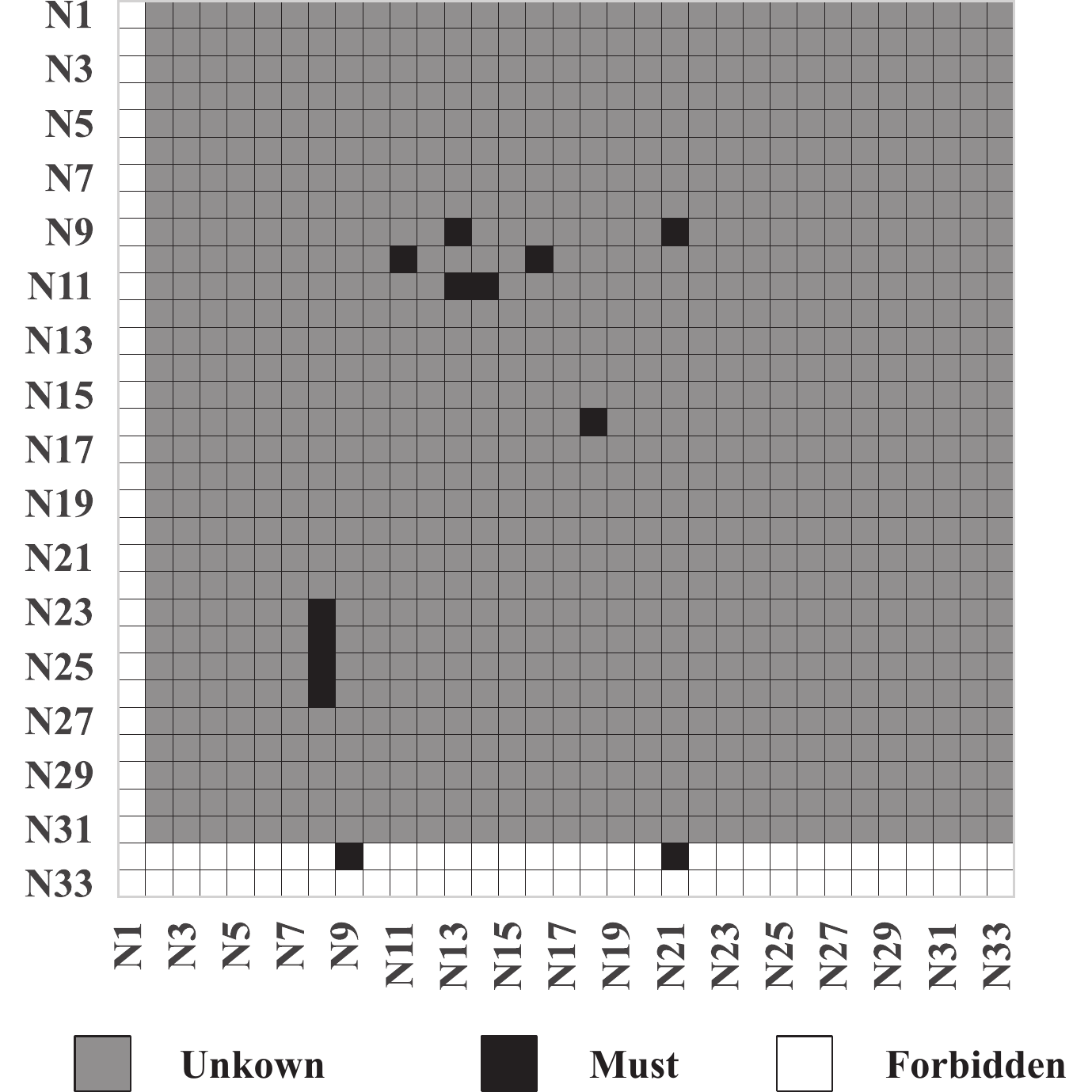}\\
    \caption{The prior knowledge matrix \emph{P}. }
\end{figure}

\begin{figure}[h]
    \centering
    \includegraphics[width=8cm,height=8cm]{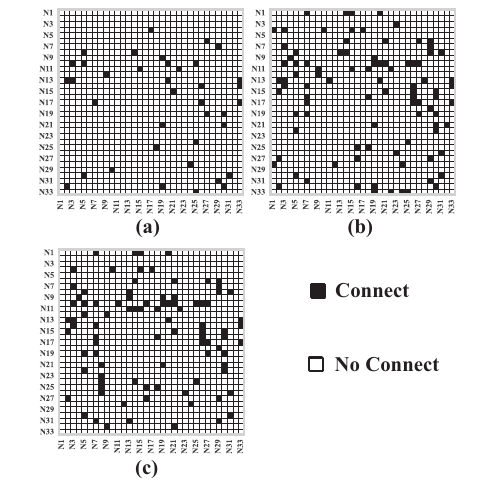}\\
    \caption{The adjacent matrix \emph{A} (a), DAG-GNN; (b), WITHTEAR (P3); (c), WITHTEAR (P4) }
\end{figure}

\begin{table}[htbp]
    \centering
    \caption{The scores of the corresponding methods}
    \begin{tabular}{@{}lll@{}}
        \toprule
        ID             & BGe      & Gaussian-BIC \\ \midrule
        DAG-GNN        & -13, 1232.0  & -52, 982.1     \\
        WITHTEARS (P3) & -85, 537.5   & -38, 237.4     \\
        WITHTEARS (P4) & -92, 276.7   & -37, 838.3     \\ \bottomrule
    \end{tabular}
\end{table}

\indent The DAG-GNN, WITHTEAR (Problem P3), and WITHTEAR (Problem P4) are given in Fig. 4. The corresponding scores are listed in Table 4. The BGe score using WITHTEAR is 29.7 $\sim$ 34.8 \% higher than that of the DAG-GNN,
and the Gaussian BIC score is 27.8 $\sim$ 28.6 \% higher than that of the using DAG-GNN.
The behavior of WITHTEAR on the score function indicates that using the MILP problem to tear the matrix rather than roughly truncate will result in a better graph that represents the data.
Note that, when solving problem P3 and P4, the BGe score and Gaussian BIC score are different, which indicates that the prior knowledge may exist bias and hence the prior should be given prudently when using WITHTEAR.

\section{Conclusions}
\indent This paper proposed the WITHTEAR to improve the problem when applying NOTEAR methods to learn DAG structure under the deep learning framework.
Firstly, the reason for the existence of infeasible solution results from the acyclic constraints adopted in the NOTEAR methods has been analyzed solved by the gradient descent method.
After that, the principle for the truncation in post-processing operation was stated from the perspective of least square perturbation analysis.
Based on this principle, the MILP models were formulated considering the tear cost and prior knowledge simultaneously in the final DAG formation.
Finally, two case studies were carried out to demonstrate the effectiveness of the WITHTEAR method comparing to NOTEAR method using DAG-GNN model as baseline.
The concentration of future work may be on the hyper-parameter before the MILP problem solving to reduce the burden of loop detection algorithm or the directly tear method face to the adjacent matrix.


%

\appendices
\section{The derivation of Matrix Derivative}
\indent The derivation of matrix derivative in Section III will be stated as follow. First, consider Eq. (1) as constraint, define $\phi$ as Eq. (A.1)shown:
\begin{align}
    \phi  = Tr({e^{A \odot A}}) = Tr(I + \sum\limits_{k = 1}^\infty {\frac{{{{(A \odot A)}^k}}}{{k!}}} ) \tag{A.1}
\end{align}
\noindent Then, the matrix derivate of trace function \cite{RN334}, \cite{IMM2012-03274} can be given as Eq. (A.2):
    \begin{align}
        d\phi & = \sum\limits_{k = 0}^\infty {\frac{{{{\left[ {{{(A \odot A)}^k}} \right]}^T}}}{{k!}}} :d(A \odot A) \notag \\
                 & = \sum\limits_{k = 0}^\infty {\frac{{{{\left[ {{{(A \odot A)}^k}} \right]}^T}}}{{k!}}} :(dA \odot A + A \odot dA) \notag \\
                 & = \sum\limits_{k = 0}^\infty {\frac{{{{\left[ {{{(A \odot A)}^k}} \right]}^T}}}{{k!}}} :2A \odot dA \tag{A.2}
    \end{align}
\noindent where the $:$ is the Frobenius $/$ trace (inner) product. Then the Hardmard product and the Frobenius product can be commuted and the Eq. (A.3) can be derived
    \begin{align}
        d\phi  = \sum\limits_{k = 0}^\infty {\frac{{{{\left[ {{{(A \odot A)}^k}} \right]}^T}}}{{k!}}}  \odot 2A:dA \tag{A.3}
    \end{align}

\indent Hence, the derivate of Eq. (1) to matrix \emph{A}$_k$ is shown as Eq. (A.4)
    \begin{align}
        \frac{{\partial \phi }}{{\partial A}} = 2A \odot \sum\limits_{k = 0}^\infty {\frac{{{{\left[ {{{(A \odot A)}^k}} \right]}^T}}}{{k!}}}  = 2A \odot {({e^{A \odot A}})^T} \tag{A.4}
    \end{align}
\indent Similarly, the matrix derivate of Eq. (2) is given as follow, define $\zeta$ as Eq. (A.5) shown.

    \begin{align}
        \zeta & = Tr[ {{{(I + \gamma A \odot A)}^d}} ] \notag \\
        & = Tr[ {\sum\limits_{k = 1}^d {C_d^k{{(I)}^{d - k}}{{(\gamma A \odot A)}^k}} } ] \notag \\
        & = Tr[ {\sum\limits_{k = 1}^d {C_d^k{{(\gamma A \odot A)}^k}} }] \tag{A.5}
    \end{align}

\noindent Then, Eq. (A.6) can be derived:

    \begin{align}
        d\zeta &  = \sum\limits_{k = 1}^d {C_d^kk{{\left[ {{{(\gamma A \odot A)}^{k-1}}} \right]}^T}} :d(A \odot A) \notag \\
        & = \sum\limits_{k = 1}^d {C_d^kk{{\left[ {{{(\gamma A \odot A)}^{k-1}}} \right]}^T}} :2A \odot dA \tag{A.6}
    \end{align}

\noindent Commute the Hardmard product and Frobenius product shown as Eq. (A.7).

    \begin{align}
        d\zeta  = 2A \odot \sum\limits_{k = 1}^d {C_d^kk{{\left[ {{{(\gamma A \odot A)}^{k-1}}} \right]}^T}} :dA \tag{A.7}
    \end{align}

\indent Finally, the matrix derivate of Eq. (2) can be given as follow:

    \begin{align}
        \frac{{\partial \zeta }}{{\partial A}} = 2A \odot \sum\limits_{k = 1}^d {C_d^kk{{\left[ {{{(\gamma A \odot A)}^{k-1}}} \right]}^T}} \tag{A.8}
    \end{align}

\renewcommand\thefigure{\thesection.\arabic{figure}}
\section{The DAG-GNN model}
\setcounter{figure}{0}
\indent In Appendix B, the DAG-GNN model will be introduced.
\begin{figure}[h]
    \centering
    \includegraphics[width=8cm,height=2.4cm]{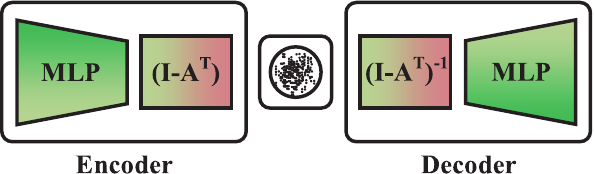}\\
    \caption{The architecture of DAG-GNN model}
    \label{Fig. B1}
\end{figure}
\noindent The architecture of DAG-GNN model is shown in Fig. B.1. The expression of encoder and decoder are given in Eq. (B.1) and (B.2) \cite{RN329}, respectively:
    \begin{align}
        \left[ {{M_Z}|\log {S_Z}} \right] = (I - {A^T})MLP(X) \tag{B.1}
        \end{align}
        \begin{align}
            \left[ {{M_X}|\log {S_X}} \right] = MLP({(I - {A^T})^{ - 1}}Z) \tag{B.2}
        \end{align}
\noindent where the \emph{M} and \emph{S}  are mean and variance respectively.
The \emph{MLP} is the multi-layer perceptron and defined as Eq. (B.3). The $W_1$, $W_2$, $b_1$, and $b_2$ in Eq. (B.3) are parameters to be learnt via gradient descent.
        \begin{align}
            MLP(X) = \operatorname{ReLU} (XW_1^T + {b_1})W_2^T + {b_2} \tag{B.3}
        \end{align}
\indent Therefore, the loss function can be given as Eq. (B.4) where the $D_{KL}$ is the Kullback-Leibler divergence, and the $p(Z)$ is the prior distribution.
        \begin{align}
            Loss & = {E_{q(Z|X)}}[\log p(X|Z)] + {D_{KL}}[q(Z|X)||p(Z)] \tag{B.4}
        \end{align}
\noindent The RHS of Eq. (B.4) can be given as follow, where $m$ is theencoder dimension, $d$ is the number of variables, $L$ is the number of variable, and $c$ is a constant:
        \begin{align}
        & {D_{KL}}[q(Z|X)||p(Z)] \notag \\
        & = \frac{1}{2}\sum\limits_{i = 1}^m {\sum\limits_{j = 1}^d {({S_Z})_{i,j}^2 + ({M_Z})_{i,j}^2 - 2\log {{({S_Z})}_{i,j}}} }  - 1 \tag{B.5}
        \end{align}
        \begin{align}
        & {E_{q(Z|X)}}[\log p(X|Z)] \notag   \\
        & \approx - \frac{1}{L}\sum\limits_{l = 1}^L {\sum\limits_{i = 1}^m {\sum\limits_{j = 1}^d {\frac{{{{({X_{i,j}} - {{(M_X^{(l)})}_{i,j}})}^2}}}{{2(S_X^{(l)})_{i,j}^2}}} } } \notag  \\
        & - \frac{1}{L}\sum\limits_{l = 1}^L {\sum\limits_{i = 1}^m {\sum\limits_{j = 1}^d {\log {{(S_X^{(l)})}_{i,j}}} } }  - c \tag{B.6}
        \end{align}
\indent To promise the acylic of \emph{A}, the constraint Eq. (2) is adopted and the optimization problem can be formulated as below :
\[\begin{array}{*{20}{c}}
{\mathop {\min }\limits_A {\text{  }}Loss = {E_{q(Z|X)}}[\log p(X|Z)] + {D_{KL}}[q(Z|X)||p(Z)]} \\
{s.t.{\text{  }}h(A) = Tr[{{(I + \gamma A \odot A)}^d}] - d = 0}
\end{array}\]

\section*{Acknowledgment}

The first author would like to thank Wenjian Du at Sun Yat-sen University and Wenshen Zhao at University of Chinese Academy of Sciences
for reviewing the mathematical derivation. He would also like to thank Le Yao and Hao Wang at Zhejiang University for
their help with the discussion of the logic of the research.

\ifCLASSOPTIONcaptionsoff
  \newpage
\fi



%
%
%
\footnotesize
\bibliographystyle{IEEEtran}
\bibliography{IEEEabrv, myref2}
\end{document}